\documentclass[letterpaper]{article}
\usepackage{aaai}
\usepackage{graphics}
\usepackage{times}
\usepackage{helvet}
\usepackage{courier}

\usepackage{algorithmic}
\usepackage{algorithm}
\usepackage{wrapfig}
\usepackage{lipsum}

\begin{document}

\title{A Framework for Complementary Companion Character Behavior in Video Games}
\author{Gavin Scott \and Foaad Khosmood \\
Computer Science Department\\
California Polytechnic State University\\
San Luis Obispo, CA\\
}
\maketitle
\begin{abstract}
\begin{quote}
We propose a game development framework capable of governing the behavior of complementary companions in a video game. A ``complementary'' action is contrasted with a mimicking action and is defined as any action by a friendly non-player character that furthers the player's strategy. This is determined through a combination of both player-action and game-state prediction processes while allowing the AI companion to experiment. We determine the location of interest for companion actions based on a dynamic set of regions customized to the individual player. A user study shows promising results; a majority of participants familiar with game design react positively to the companion behavior, stating that they would consider using the framework in future games themselves.
\end{quote}
\end{abstract}



\subsection{Introduction}
\noindent As video games evolve, they become more and more complex; many best selling AAA (big-budget) game titles now contain complex three-dimensional worlds with rich stories and hundreds of hours of unique content. With this complexity has come the need to better utilize artificial intelligence (AI) in games. Some games use AI to adapt the game mechanics to the style of a particular player \cite{denisova2015dynamic}. Others use it to make the characters in the game more intelligent, causing them to behave more naturally. Part of player immersion in role playing games (RPG) or real-time strategy games (RTS), is the quality of the AI-controlled non-player characters (NPC), and their ability to interact with both each other and the player in intuitive ways \cite{diaz2017npc}. \par

The purpose of a companion character in a video game is often to assist the player character in their next objective, which in some cases means performing complementary actions that help the player achieve their goal more easily than they could on their own. This is rarely successful in practice, and the companion is often relegated to a beast of burden, or a lackey that follows scripted behavior \cite{gemine2012imitative}. Poorly-designed companions can diminish the overall game experience; a teammate should work with the player and change their strategy to suit their needs, making the game more interesting, engaging, and enjoyable. Following a small, fixed set of decisions can work for games with a linear narrative where the player's overall strategy is predetermined, but more complex games require more advanced behavior. It is not practical to code anticipatory behavior ahead of time to satisfy all situations. What is needed is an expressive, robust and flexible decision-making system for the companions.\par

\subsection{Lord of Towers}
\begin{figure}[H]
  \includegraphics{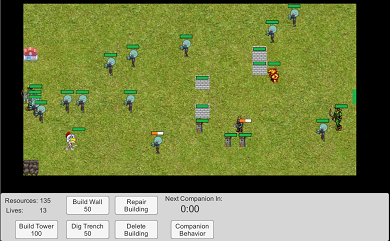}
  \caption{A screen-shot of Lord of Towers gameplay}
  \label{fig:lot}
\end{figure}

``Lord of Towers'' is an unpublished top-down tower defense game developed in Unity to test the ``MimickA'' \cite{angevine2016mimica} framework from previous work. The player controls a character that can move around the map, using limited resources to construct and repair defensive buildings to defend against an infinite wave of enemies all attempting to reach the players base. There are various types of enemies, each with slightly varied attributes (special abilities, varying speed and health, etc.), and they enter the map from the right side, moving to attack the player's base on the left. The player's goal is to survive as long as possible; the game ends when a certain number of enemies reach the player's tower. Figure \ref{fig:lot} shows a screen-shot of Lord of Towers.

The game also includes a companion character that enters the map after a few minutes of gameplay. Modifying the behavior of this companion to was the central focus of this project. In previous work, the ``MimickA'' system controlled the companion character which used machine learning to mimic the player's behavior by using training data produced live by the human player. In this work, the companion uses this same data to attempt to perform complementary (rather than mimicking) actions, based on criteria that will be discussed in later sections. An important change to gameplay that was introduced for this project was an intermittent blacking-out of the screen; this negatively impacted gameplay and the results of the user study, but was necessary to implement our desired approach within the limits of the system in which Lord of Towers was designed.

\subsection{Overview of our Solution}
This project extends the framework ``MimickA'' \cite{angevine2016mimica} to include dynamic complementary decision-making for a companion AI. MimickA focused on learning player behavior and having the companion make similar decisions, given similar states of the world. We modify this framework to find actions that complement rather than mimic the player's strategy. Determining to what degree an action is complementary to a strategy is difficult to quantify, so we conduct a user study to gauge our system's success. \par

\begin{figure}[H]
  \includegraphics{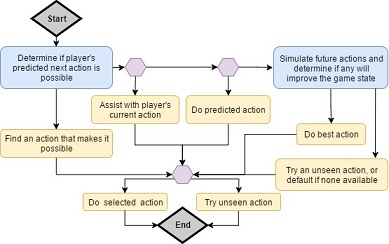}
  \caption{A simplified version of the decision-making process}
  \label{fig:micro}
\end{figure}

Specifically, we develop a complementary decision-making process that combines evaluating the player's current and past actions, predicting their likely next action and future game states, and experimenting with available actions not yet performed by the player to increase the companion's range of possible behaviors. This allows the actions the companions take to be tailored to the behavior of the player without the need for scripted behaviors or off-line training. We also implement a dynamic region system for determining where actions should be taken. Figure \ref{fig:micro} shows a basic overview of the companion decision-making process. The small hexagons represent decisions that are made based on probabilities set by the developer, allowing each game to have customized behavior.\par

\section{Related Work}
Much of the relevant literature falls into two groups: player strategy prediction and adaptive AI approaches.

\subsection{Player Strategy Prediction}
Often, the goal of player modeling is not to predict specific actions and when the player will perform them, but instead using human player action traces taken from previous play sessions to predict how the current one will behave. This can be used to dynamically adjust the difficulty of the game to keep the player engaged without becoming frustrated \cite{denisova2015dynamic}, or to make predictions such as how long it will take the player to finish the game \cite{mahlmann2010predicting}.\par

Some methods for identifying the player strategy rely on collecting a large number of game traces offline to find similar strategies to a current situation \cite{mahlmann2010predicting,gemine2012imitative}. A simple example of this technique is shown in an experiment where researchers attempted to make a small set of predictions about a player's performance in ``Tomb Raider: Underworld'' \cite{mahlmann2010predicting}. In-game data was collected from a large number of players, and as a new person played the game their strategy was analyzed and the data was used to predict whether or not they would finish the game, and how long it would take them to do so. Even with only a small set of relatively simple factors to predict, their accuracy was fairly low and the authors concluded their techniques would not be accurate enough for real-time game adaptation but could be useful as a source of feedback on the game design. \par

\subsection{Adaptive A.I.}
The goal of adaptive AI is to alter the behavior of computer controlled characters, rather than the gameplay, for each individual player. Adaptive game AI can be applied to a player's teammates \cite{feng2016npc,karlsson2015,guckelsberger2016}, or applied to the enemy AI \cite{silva2015dynamic}. In either case, the computer-controlled character learns to react to each individual player differently; this can make the relationship between the player and the team more rewarding, or be applied to enemies to make the game more challenging. We applied these techniques to the companion AI to develop a teammate that is more effective than one with statically-defined behavior. \par

\section{Defining ``Complementary"}

To create a working definition of a ``complementary" action, we investigated psychology research regarding complementary \cite{complementaryactions} and ``pro-social'' behavior \cite{pyschhandbook}, as well as ``joint-action'' behavior \cite{knoblich20113}. Based on these sources, complementary behaviors can be defined as multiple agents coordinating different actions to further a common goal \cite{complementaryactions}. Distinctions have also been made between ``planned'' and ``emergent'' types of agent coordination \cite{knoblich20113}. \par

Complementary behavior is not necessarily imitative \cite{complementaryactions}. An agent performing an identical action to their teammate may further the team's overall goal, but it is quite possible that the goal would be better served by a different action. In the context of a video game, an imitative companion is limited in their behavior; if the player has never performed an action the companion will avoid it as well, neglecting potentially useful assistance. At the other extreme, if the companion completely ignores the player's set of actions they may repeatedly do something that the player was intentionally avoiding, negatively affecting their strategy. \par

Imitative agents are also limited by their lack of understanding of what result an action will have. They blindly perform actions that they believe the player would perform, without understanding the effects that those behaviors have on the game. If a game has an overall goal (increasing the score, etc.), the companion that cannot predict the outcomes of their actions may unknowingly hinder progress towards that goal. 

It is also necessary for a complementary agent to guess the future actions of its teammates as well as the effects that its own actions might have on the rest of the game \cite{complementaryactions}. By predicting what the player plans on doing next, the companion can consider that action the player's ``immediate goal'' and attempt to facilitate it. Without the ability to predict the outcome of a potential action, an agent would be unable to judge how to best enable the player to pursue their chosen strategy. \par

To synthesize these requirements, we present a set of guidelines for determining an ideal companion action. We consider an action to be complementary if these rules are followed:

\begin{itemize}
    \setlength\itemsep{0em}
    \item The proposed action should remove obstructions to the player's next predicted action, if possible. This prioritizes the individual player's goal over the overall goal of the game, limiting player frustration as much as possible by streamlining their strategy.
    \item If there are no obstructions to the player's next action or they cannot be removed, the companion action will attempt to improve the overall state of the game in their favor, as determined by developer-defined metrics (score, health, resources, etc.).
    \item A potential action should be avoided if it jeopardizes the player's next predicted action for the sake of other perceived benefits, like increased score. This further re-enforces the criteria to prioritize the individual player's strategy.
    \item The distribution of actions the companion takes should be similar to the player's, but they should not be restricted to only performing actions that the player has previously performed. This allows room for limited experimentation with unseen behavior, but prevents them from performing an overabundance of new actions that the player may have been intentionally avoiding.
\end{itemize}
\section{System Design: Decision Making}

This section provides a high-level overview of the decision-making process used by the companion to identify actions that are complementary to the player's immediate and overall goal. This process is based on the current game state and the previously-defined guidelines for choosing a complementary action. A simplified flowchart visualizing this process is included in Figure \ref{fig:micro}. \par

\subsection{Main Process} \label{main_seq}

First, the system predicts the action that the player is likely to take after they finish their current action (called ``next action''), based on previous data. How this prediction is made is explained in the ``Player Prediction" section. The companion checks whether or not the predicted action is possible in the current game state, based on a framework-defined method implemented by the game developer. If the predicted action is impossible, the system searches for an action that results in a game state where the next action can be performed, by predicting future game states that would result from each possible action (this process is explained in the ``State Prediction" section). \par 

If the player's next action is possible, then the companion checks if they are able to assist with the player's current action, and may choose to help (based on a stochastic decision model with a probability set by the developer during the framework's configuration). This is included as a possibility because helping the player finish their current action more quickly definitively helps further their immediate strategy. \par

If the companion can't help the player with their current action, they randomly choose one of two paths (probabilities are configurable). First of the two options is for the companion to begin performing the player's predicted next action. Assuming the player's strategy consists of a series of actions, this behavior allows the actions to be performed in parallel, speeding up the process. If the second branch is taken, the companion tries to find an action that results in the ``best'' game state. This metric is not player-specific, and is an overall goal that is set by the developer (increase score, keep player's health high, etc.). This process involves predicting the outcome of potential actions available and ranking their outcomes. Actions that prevent the player from performing their predicted next action are removed to avoid jeopardizing the player's strategy, and the highest-scoring remaining action is performed. \par

If all possible actions are rejected, the companion takes a last resort action of looking for any action that is possible but that the player has not yet performed. If no such actions exist (the player has performed every action at least once), the companion resorts to a default behavior that can be defined by the developer. \par

\subsection{Final Decision Sequence} \label{final_seq}

After an action is chosen, if there are actions available to the companion that the player has never performed, there is a chance that an unseen action will be randomly chosen instead of the companion's initial decision. This step forces the companion to occasionally experiment, ideally exposing the player to strategies that they had not considered. \par
\section{Implementation}

\subsection{Framework}
To maximize the usefulness of our framework, each feature is implemented to be as configurable as possible. The game developer can control and tune as many of the features as possible; every stochastic decision probability, which classifiers to use and how often they're trained, what features to use for state prediction, etc. \par

\subsection{Dynamic Region System} \label{regions}

Determining which action to take is only part of the companion's decision; the framework must also determine where the actions will take place. Lord of Towers was originally implemented with a fixed-size map split into six equal sectors. The large and static size of these areas posed a problem by limiting the granularity of the companion's choices. In ``Lord of Towers" in particular, players have a strong tendency to perform the majority of their actions in the few sectors near their base, effectively reducing the usefulness of the companion's decision and offloading a large amount of the decision-making process to the game developer. \par

\begin{figure}
  \centerline{
    \includegraphics{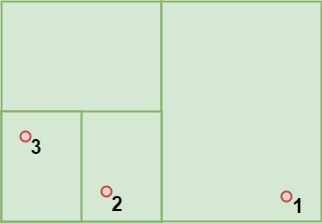}
  }
  \caption{An example of the region system, showing the regions after the player's first three actions.}
  \label{fig:region_example_1}
\end{figure}

An alternative, dynamic region system was adopted to address this issue and give more power to the framework while still leaving a specific location choice to the developer. In this system, the map is initially set to one large region. Every action the player takes creates a new region by finding the region in which the action took place and splitting it along it's shortest axis. Each of the player's new actions are stored with their global coordinates; action-region pairs that are returned later contain the region found by looking up that coordinate in the most up-to-date set of regions, not the regions that were present when the action was taken. Figure \ref{fig:region_example_1} shows an example of how a map's regions might look after three events are taken in sequence.

This technique is very similar to a simplified Binary Space Partitioning (BSP) tree, the primary differences being that this technique can be updated with each new action in constant time, and region sizes are less varied \cite{bspsource}.\par

These regions provide location data that is more specific than the previous large sectors, while still allowing the game designer the leeway to decide where within a region an action should take place (exact coordinates are not returned). It also lends itself well to choosing a random region, which happens when previously-unseen actions are tried; areas of the map with a higher density of player actions will have more companion actions, but the companions still has a chance to experiment and try a region that the player has left relatively unexplored. \par

\subsection{Player Action Prediction} \label{player_prediction}

Prediction of the player's next action is handled using a classifier trained on the player's previous data, requiring the game to not introduce the companion until enough data has been recorded to make this prediction. The type of this classifier (as well as when to introduce the companion) can be set by the developer; a decision tree was used in Lord of Towers as the main classifier. The data is stored as an ordered list of pairs of game state vectors and actions with a global coordinate attached that indicates where on the map the player took the action. When the classifier is trained, a secondary data set is built from the player's data and used to fit the classifier. By default the entire game state vector is used but the game developer has the option to choose to manipulate this data if using the full feature vector is not desired. A method is provided to try multiple trials of both different feature vector configurations as well as different classifiers on a set of player data, returning the method that had the highest average accuracy. \par

Once the classifier is trained, the current game state vector can be used to predict an action-location pair. Rather than return the location to the developer as a single point in global coordinates, the region containing the point is returned based on the current global set of regions. This is then used later in the companion's decision-making process. \par

\subsection{Game-State Prediction} \label{state_prediction}

Multiple parts of the companion's decision-making process require the companion to predict how each possible action will affect future game states, choosing an action based on which results in the best foreseen outcome. To facilitate the comparison of the predicted game states, the abstract class representing a state feature vector specifies a method that returns a numerical score; once a mapping of possible events to predicted vectors is created, the action with the highest-scoring vector is chosen. 

To build the mapping of action to vector, we chose to have the companion actually perform each action for a given number of frames to allow the effects of the action to be recorded. Originally we used traditional machine learning techniques to predict the game state given the action, but this required too much data and was too inaccurate. Instead, we decide to learn by simulating the actual world after the proposed action is performed, then evaluating the metrics of success (score, life, etc.) as defined by the designer. We implement saving and loading functionality so the game state can be preserved before the prediction and reset before each new action. Pseudo-code describing the full state prediction process is included below in algorithm \ref{alg:dec_alg}.\par

\begin{algorithm}[H]
\caption{Game-State Prediction}
\begin{algorithmic}[1]
    \STATE origState, origGameSpeed $\leftarrow$ saveGameState()
    \STATE setGameSpeed(maxSpeed)
    \STATE actionScores $\leftarrow$ emptyHashMap
    \FOR{action, region in getSeenPairs(actions, regions)}
        \IF {isPossible(origState, action, region)}
            \STATE newState $\leftarrow$ performAction(action, region, time)
            \STATE{actionScores[action, region] $\leftarrow$ score(newState)}
            \STATE resetGameState(origState)
        \ENDIF
    \ENDFOR
    \STATE setGameSpeed(origGameSpeed)
    \STATE bestAction, bestRegion $\leftarrow$ findMaxScore(actionScores)
    \RETURN bestAction, bestRegion
\end{algorithmic}
\label{alg:dec_alg}
\end{algorithm}

After the original game state and speed are saved, the game is sped up to reduce the processing time (since the player is not interacting with the game during this period, playability is not an issue). All action-region pairs in the current player's history are iterated over (see ``getSeenPairs'' in line 4), and actions that are currently possible in their paired region are simulated for a specific time set by the designer (see ``performAction'' in line 6). Each action-region pair is then given a score based on the new game state after the simulation ends, calculated using a developer-defined method (see ``score" in line 7). Because the game state has been changed, it is then reset to its initial value so the next simulation can be run with the same starting parameters. Once all action-pairs have been evaluated, the game is reset to its initial configuration once more and the action with the highest-scoring end-states is returned (see lines 12 and 13). \par
\section{Experimental Design}
We conducted a 25-user study of undergraduate college students in a video game design class. Subjects played Lord of Towers with our companion for twenty minutes total, repeating the game if they lost. To reduce user-bias, the subjects were not aware that the companion was the focus of the project. The survey given afterwards asked generic questions first before specifically asking about the behavior of the companion. The questions on the survey are focused on the topics below. The full text of these questions is included in the "Selected Questions" section. \par

\begin{itemize}
    \setlength\itemsep{0em}
    \item How much they enjoyed the game, and the familiarity with Tower Defense games.
    \item The general strengths and weaknesses of the game.
    \item The behavior of the companion.
    \item The potential usefulness of the framework to developers (they were given a description of the framework).
    \item The effect the blacking-out of the screen during the prediction process (an implementation prototype side-effect) on the game play and their other opinions.
\end{itemize}
\section{Results \& Evaluation}

\begin{figure}
  \centerline{
    \includegraphics{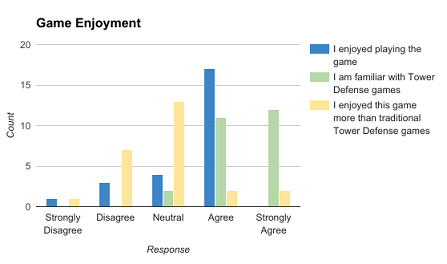}
  }
  \caption{A summary of responses to questions regarding participant enjoyment of the game}
  \label{fig:enjoyment}
\end{figure}

Overall, the results of the user study are positive. The majority of the participants enjoyed the game, although not more than most tower defense games. A summary of the participants responses to these questions is shown in figure \ref{fig:enjoyment}. To reduce player bias, the participants were asked about the game's strengths and weaknesses while they were still unaware of the purpose of the project, before specifically asking about the companion. Many of these responses were unrelated to companion AI (complementing the gameplay, etc.), but seven of the twenty-five participants singled out the companion as a strength, writing comments such as ``the companion AI was pretty good and helpful in most cases" and ``the companion would actively help you." This is encouraging, showing that the companion behavior was of a high-enough quality to be noteworthy before the users were aware that it was the focus of the project. Two participants even noted that the companion appeared to act how they would expect a human teammate to behave, saying ``[it behaved how] I would expect from a friend playing the game (like a slower version of a human player)" and ``It felt almost like half of a player. His actions were as beneficial as my own (to some extent) and did not feel like a pet or random NPC that you'd usually ignore." \par

\begin{figure}
  \centerline{
    \includegraphics{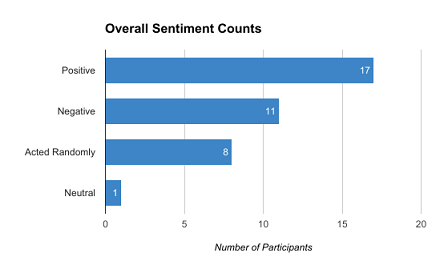}
  }
  \caption{A summary of answer sentiments to free-response questions regarding the behavior of the companion AI}
  \label{fig:sentiment}
\end{figure}

A number of questions were then asked specifically about the companion behavior, the results of which are summarized in Figure \ref{fig:sentiment}, which shows a count of the number of individual participants organized by the estimated sentiment of their comments. The statements were categorized into one or more of the four categories based on the following criteria: any comment that specifically mentioned the companion learning from the player's strategy was recorded as positive, and complaints about the companion were considered negative. A number of people also commented that the companion seemed to act randomly; this was a common response in the original ``MimickA" paper, so it was included as a category here as well. We attribute this to the relatively small set of actions that the companion can perform which might have made it hard to determine its motives. Some players also mentioned that they were unable to play too close attention to the companion because they were absorbed with the rest of the game; responses like this were considered neutral. \par

While most participants thought that the companion was helpful, some viewed their actions as wasting resources or not doing the actions in a helpful way. We assume that the strong differences in opinion may be a result of how the companion's behavior differs player to player. It is theoretically possible that certain play-styles, at least in Lord of Towers, were better suited to the framework than others and led to more useful actions. This could be remedied by incorporating player feedback in-game or having a base level of scripted behavior that constrains the actions returned by the framework. It may also be that some players prefer to be in absolute control of their in-game resources, and any unauthorized spending would be an annoyance.  \par

\begin{figure}
  \centerline{
    \includegraphics{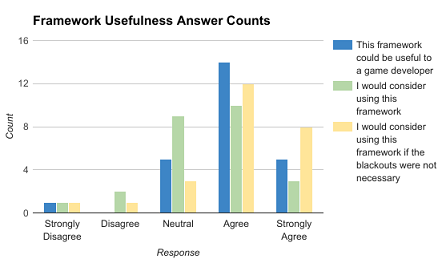}
  }
  \caption{A summary of responses to questions regarding the usefulness of the framework}
  \label{fig:usefulness}
\end{figure}

After recording the answers to these questions, the users were given a description of the project and a summary of the decision-making process. They then answered questions about its perceived usefulness to them as game developers; results are summarized in Figure \ref{fig:usefulness}. Most participants reacted favorably, indicating that the framework could indeed be useful in a general sense. They also all indicated that the largest weakness of the system was the blackouts during the state-prediction process, and that the framework without that limitation would be more appealing. \par

\begin{figure}
  \centerline{
    \includegraphics{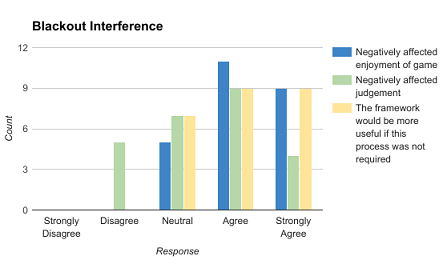}
  }
  \caption{A summary of responses to questions regarding the effect of the state-prediction blackouts on participant impressions of the game}
  \label{fig:blackout}
\end{figure}

Many of the participant responses mentioned that the blackouts during the companion's state-prediction process were a significant problem in terms of user experience. This was expected and questions were included to gauge this effect, the answers to which are summarized in Figure \ref{fig:blackout}. We expect that based on these results, the responses to the other questions would have improved in at least some cases. It would be interesting to re-run the user study after modifying the system to run the state-prediction process in a separate thread. \par

\section{Conclusion}
We proposed a game-development framework for adding novel complementary companion behavior to a tower-defense video game. A complementary action is defined as one that furthers the individual player's strategy, and is determined through a combination of player-action and game-state prediction. We conduct a user study where participants play a game containing companions using our framework, and receive promising results: many participants single out the companion without prompting, and a majority of participants agree that the framework would be useful to them as developers. We would like to continue this project to further improve it and remove some technical limitations we encountered with this version. \par
\section{Future Work} \label{future}

The largest roadblock towards this system's usefulness is the single-thread limitation imposed by Unity that forces the game-play to halt when a companion uses state-prediction to choose an action. To fix this, the system should be migrated to a new environment. This could allow for multiple companions to be present at once, hopefully allowing for the exploration of complementary team dynamics. \par

The choice of the probabilities for the randomized aspects of the companion's decisions would also potentially benefit from further investigation. The current values are static and chosen subjectively. It would be possible to allow these values to change dynamically during game-play as the companion learned from its decisions. Alternatively some method of player feedback could be introduced allowing them to prioritize certain actions over others, and the probability could be modified accordingly. This would support some of the suggestions received during the user study.\par

The state-prediction process could also be extended to have the companion try sequences of actions rather than one at a time. This would result in a longer but less frequent process, action sequences that go particularly well together (creating a building and then repairing it to full health, for example) could be identified and paired together into a new ``single" action.\par
\section{Selected Questions}
\begin{enumerate}
    \itemsep0em
    \item What would you say the game's strengths were? (asked again for weaknesses) [free form text]
    \item Did you notice anything noteworthy about the AI in the game? [free form text]
    \item How would you describe the behavior of the companion? [free form text]
    \item The change in gameplay (blacking out the screen, etc.) during the companion decision-making process negatively affected my enjoyment of the game. [Likert scale: ``strongly agree'' to ``strongly disagree'']
    \item The change in gameplay negatively affected how well I could judge the quality of companion decisions. [Likert scale: ``strongly agree'' to ``strongly disagree'']
\end{enumerate}

\bibliographystyle{aaai}
\bibliography{main.bib}

\end{document}